\def\year{2022}\relax
\documentclass[letterpaper]{article} % DO NOT CHANGE THIS
\usepackage{aaai22}  % DO NOT CHANGE THIS
\usepackage{times}  % DO NOT CHANGE THIS
\usepackage{helvet}  % DO NOT CHANGE THIS
\usepackage{courier}  % DO NOT CHANGE THIS
\usepackage[hyphens]{url}  % DO NOT CHANGE THIS
\usepackage{graphicx} % DO NOT CHANGE THIS
\usepackage{natbib}  % DO NOT CHANGE THIS AND DO NOT ADD ANY OPTIONS TO IT
\usepackage{caption} % DO NOT CHANGE THIS AND DO NOT ADD ANY OPTIONS TO IT
\DeclareCaptionStyle{ruled}{labelfont=normalfont,labelsep=colon,strut=off} % DO NOT CHANGE THIS
\usepackage{amsmath}
\usepackage{amsfonts}
\usepackage{array}
\usepackage{bm}
\usepackage{booktabs}
\usepackage{multirow}
\usepackage{tikz}

\graphicspath{ {./images/} }
\usetikzlibrary{bayesnet, calc, positioning}

\title{Multimodal Adversarially Learned Inference with Factorized Discriminators}
\author {
    Wenxue Chen,\textsuperscript{\rm 1}
    Jianke Zhu\textsuperscript{\rm 1,2\thanks{corresponding authors}}
}
\affiliations {
    \textsuperscript{\rm 1} Zhejiang University \\
    \textsuperscript{\rm 2} Alibaba-Zhejiang University Joint Institute of Frontier Technologies\\
    \{wxchern, jkzhu\}@zju.edu.cn
}
\begin{document}

\maketitle

\begin{abstract}
    Learning from multimodal data is an important research topic in machine learning, which has the potential to obtain better representations. In this work, we propose a novel approach to generative modeling of multimodal data based on generative adversarial networks. To learn a coherent multimodal generative model, we show that it is necessary to align different encoder distributions with the joint decoder distribution simultaneously. To this end, we construct a specific form of the discriminator to enable our model to utilize data efficiently, which can be trained constrastively. By taking advantage of contrastive learning through factorizing the discriminator, we train our model on unimodal data. We have conducted experiments on the benchmark datasets, whose promising results show that our proposed approach outperforms the-state-of-the-art methods on a variety of metrics. The source code will be made publicly available.
\end{abstract}

\section{Introduction}
\noindent Real-world data usually have several heterogeneous modalities. Co-occurrence of different modalities provides the extra supervision signals, since the different modalities of data are usually associated by the underlying common factors. By exploiting such commonality among different modalities, the representations learned from the multimodal data are promised to be more robust and generalizable across different modalities. Existing works have shown the potential of the representations learned from the multimodal data~\citep{tsai2018learning,shi2021relating}.
%(TODO: examples of both supervised multimodal learning and unsupervised multimodal learning here....)

Supervised representation learning requires the labeled data, which are typically scarce comparing to the vast amount of unlabeled data. Recently, some progress on generative modeling has been made in the multimodal setting and demonstrates the power of the representations learned from the multimodal generative models~\citep{DBLP:conf/iclr/SuzukiNM17,Wu2018MultimodalLearning,shi2019variational,NEURIPS2020_43bb733c}. In this work, we focus our attention on generative modeling of the multimodal data. 

Generative models offer an elegant way to learn from the data without labels. Variational auto-encoders (VAEs)~\citep{DBLP:journals/corr/KingmaW13} and generative adversarial networks (GANs)~\citep{NIPS2014_5ca3e9b1} are two well-known representative deep generative models. In the original formulation, GANs are not able to perform inference like VAEs, the adversarially learned inference (ALI) and bidirectional GAN (BiGAN) \citep{DBLP:conf/iclr/DonahueKD17,DBLP:conf/iclr/DumoulinBPLAMC17} proposed an extension by training the inference model and the generative model together using the adversarial process.

Recent approaches~\citep{DBLP:conf/iclr/SuzukiNM17,shi2019variational,NEURIPS2020_43bb733c} have extended the VAEs to the multimodal setting, where learning of the model corresponds to maximize the evidence variational lower bound (ELBO) on the marginal log-likelihood of multimodal data. However, they are not able to utilize the unimodal data efficiently, as combining unimodal ELBOs and multimodal ELBOs naively does not yield a valid lower bound on the marginal log-likelihood of multimodal data anymore, which may even hurt performance. \citet{shi2021relating} propose a constrastive version of multimodal ELBO to improve data-efficiency, where the model can make use of the unpaired unimodal data. However, their optimization objective is derived heuristically and empirically weighted to avoid degenerate solutions, where the model only learns to generate the random noise.

To address the above limitations, in this paper, we propose a novel GAN-based multimodal generative model to make use of unimodal data more efficiently. We show that it is necessary to align the different encoder distributions with the joint decoder distribution, simultaneously. To align multiple distributions, we resort to adversarial learning. Specifically, training the discriminators in the D-step of GANs can be regarded as estimating the probability density ratio. This perspective of GANs enables us to factorize the discriminators and compute a wide range of divergences. By factorizing discriminators, we can train discriminators on unimodal data, which improves the data-efficiency. By taking advantages of the flexibility of GANs, we propose a principled way to combine the unimodal and multimodal training under the same framework.

From the above all, the contributions of our work are summarized in threefold as below:
\begin{itemize}
    \item We show that the different encoder distributions and decoder distribution need to be aligned simultaneously to learn a coherent multimodal generative model.
    \item We construct a specific form for the discriminator that is able to take advantage of unimodal data and contrastive learning.
    \item Our proposed approach outperforms the-state-of-the-art methods on a variety of metrics.
\end{itemize}

The rest of this paper is organized as follows. We summarize the desiderata of multimodal generative modeling in Section~\ref{sec:background}. We propose the optimization objective to address the desiderata in Section~\ref{sec:mmali}. We show how to factorize the joint discriminator to enable contrastive learning in Section~\ref{sec:dis_factor} . In Section~\ref{sec:latent_factor}, we show that it is necessary to factorize the latent space. The experiment results are presented in Section~\ref{sec:exps}. We discuss some related works in Section~\ref{sec:related}. Finally, Section~\ref{sec:cons} gives the conclusions.

\section{Backgroud}
\label{sec:background}

To equip GANs with inference mechanism, \citet{DBLP:conf/iclr/DonahueKD17,DBLP:conf/iclr/DumoulinBPLAMC17} proposed to align the encoder distribution $q(\bm{x}, \bm{z}) = q(\bm{x}) q(\bm{z} | \bm{x})$ and the decoder distribution $p(\bm{x}, \bm{z}) = p(\bm{z}) p(\bm{x} | \bm{z})$ using an adversarial process:
\begin{align}
\begin{split}
  \label{eq:unimodal_value_equation}
  \min_G \max_D V(D, G) & =
  \mathbb{E}_{q(\bm{x})}[\log(D(\bm{x}, G_{\bm{z}}(\bm{x})))] \\
    & \quad + \mathbb{E}_{p(\bm{z})}[\log(1 - D(G_{\bm{x}}(\bm{z}), \bm{z}))]
\end{split}
\end{align}
where $q(\bm{x})$ is the data distribution, and $p(\bm{z})$ is the prior imposed on the latent code. $G_{\bm{z}}(\bm{x})$ and $G_{\bm{x}}(\bm{z})$ are the encoder and decoder corresponding to conditional distributions $q(\bm{z} | \bm{x})$ and $p(\bm{x} | \bm{z})$, respectively. The discriminator $D$ takes both $\bm{x}$ and $\bm{z}$ as input, then determines if the pair comes from the encoder distribution or the decoder distribution. 

By training an encoder together, the inference of the generative model can be performed via $p(\bm{z} | \bm{x}) = q(\bm{z} | \bm{x})$. When the encoder network $G_{\bm{z}}(\bm{x})$ and the decoder network $G_{\bm{x}}(\bm{z})$ are deterministic mappings and the global equilibrium of the game is reached, i.e., $q(\bm{x}, \bm{z})=p(\bm{x}, \bm{z})$, they invert each other almost everywhere (i.e., $\bm{x} = G_{\bm{x}}(G_{\bm{z}}(\bm{x}))$ and $\bm{z} = G_{\bm{z}}(G_{\bm{x}}(\bm{z}))$) even though the models are trained without the explicit reconstruction loss. 

Aligning encoder and decoder distributions does not guarantee faithful reconstruction in the stochastic model due to non-identifiability of the model. Also, the global equilibrium might never be reached in practice. \citet{NIPS2017_ade55409} illustrated that we can still achieve autoencoding property through stochastic mappings by adding the conditional entropy regularization terms to the original ALI/BiGAN objective.
% regularize the conditional entropy $H_{q}(\bm{x} | \bm{z})$ under the encoder distribution $q(\bm{z} | \bm{x})$.

% recent works have focused on variational auto-encoders in the multimodal setting.

In the unimodal setting, only generation $p(\bm{x} | \bm{z})$ and inference $q(\bm{z} | \bm{x})$ are considered.
However, the situation becomes more complicated in the multimodal setting when we try to model the multimodal data distribution $q(\bm{X})$ with the sample $\bm{X} = \{\bm{x}_i\}_{i=1}^M$ having $M$ modalities.
As in the unimodal case, we intend to generate multimodal data $p(\bm{X} | \bm{z})$ as well as perform inference on fully observed multimodal data $q(\bm{z} | \bm{X})$. Since different modalities may be missed during inference, we want to perform inference given only partial observation $q(\bm{z} | \bm{X}_K)$, where $\bm{X}_K$ is a subset of $K$ modalities observed. Furthermore we aim to generate other unobserved modalities conditioned on partial observation.
\citet{shi2019variational} have recently proposed desiderata for multimodal learning.
We thus summarize the desiderata here for our multimodal generative model as follows:
% Recently they have proposed to satisify following desiderata
\begin{itemize}
    \item \textbf{Desideratum 1. Latent Factorization} The latent code learned by the model can be factorized into the modality-specific code and the code shared across modalities.
    \item \textbf{Desideratum 2. Coherent Joint Generation} The samples generated by the model should follow the joint distribution of the multimodal data.
    \item \textbf{Desideratum 3. Coherent Cross Generation} Given one set of modalities, the model can generate other set of modalities that follows the conditional distribution of the multimodal data.
    \item \textbf{Desideratum 4. Synergy} The model should benefit from observing more modalities.
\end{itemize}

\section{Proposed Method}
%In the following sections, we show that by satisifying above desiderata, we can further factorize the 

\subsection{Multimodal Adversarially Learned Inference}
\label{sec:mmali}
In the multimodal setting, the observation is the multimodal data, from which we are interested in learning the shared latent representation for all modalities.

Given the multimodal data $\bm{X} = \{\bm{x}_i\}_{i=1}^M$ with $M$ modalities and the latent code $\bm{z}$, we consider the following $M$ different unimodal encoder distributions:
\begin{align}
    q_i(\bm{X}, \bm{z}) = q(\bm{x}_1, \dots, \bm{x}_M) q(\bm{z}|\bm{x}_i)
\end{align}
where $q(\bm{X})$ is the joint distribution of the multimodal data. $q(\bm{z} | \bm{x}_i)$ is the conditional distribution corresponding to the encoder $G_{\bm{z}}(\bm{x}_i)$ that maps the modality $\bm{x}_i$ to the latent code $\bm{z}$. The samples from distribution $q_i(\bm{X}, \bm{z})$ are generated by the following steps. Firstly, we take a multimodal data sample $\bm{X}$ from the data distribution $q(\bm{X})$, and then generate the latent code $\bm{z}$ using only one modality $\bm{z} = G_{\bm{z}}(\bm{x}_i)$.

With the assumption that different modalities are conditionally independent given the code $\bm{z}$, we consider the following decoder distribution:
\begin{align}
    p(\bm{X}, \bm{z}) &= p(\bm{z})p(\bm{x}_1 | \bm{z}) \dots p(\bm{x}_M | \bm{z})
\end{align}
%$q(\bm{x}_i) q(\bm{z} | \bm{x}_i) = p(\bm{z}) p(\bm{x}_i | \bm{z})$ which means we can generate other modalities conditioned on the modality $\bm{x}_i$ as we then have
where $p(\bm{z})$ is the prior distribution imposed on $\bm{z}$, and $p(\bm{x}_i | \bm{z})$ is the conditional distribution corresponding to the decoder $G_{\bm{x}_i}(\bm{z})$ that maps the latent code $\bm{z}$ to modality $\bm{x}_i$. 

If we align the distribution $p(\bm{X}, \bm{z})$ with one of the encoder distributions $q_i(\bm{X}, \bm{z})$, then we can make sure that: (i) the generative model learns coherent generation, i.e., $p(\bm{X}) = q(\bm{X})$, which addresses desideratum 2 (coherent joint generation); (ii) we can perform inference on modality $\bm{x}_i$ as $p(\bm{z} | \bm{x}_i) = q(\bm{z} | \bm{x}_i)$ and cross generation from $\bm{x}_i$ to other modalities is coherent as $q(\bm{x}_i) q(\bm{z} | \bm{x}_i) \prod_{j \neq i}^M p(\bm{x}_j | \bm{z}) = p(\bm{z})\prod_i^M p(\bm{x}_i | \bm{z})$. It is necessary to align the decoder distribution $p(\bm{X}, \bm{z})$ and every unimodal encoder distribution $q_i(\bm{X}, \bm{z})$ simultaneously, if we want to condition on arbitrary modality. Furthermore, if we want to condition on multiple modalities at the same time, we can introduce multimodal encoder distributions $q(\bm{X})q(\bm{z} | \bm{X}_K)$, where $\bm{X}_K$ is a subset of $\bm{X}$ with $K$ modalities, and align it with the decoder distribution as well. To fully address the desideratum 3 (coherent cross generation), we need to align all the unimodal and multimodal encoder distributions with the decoder distribution, simultaneously.

% We do not need to introduce extra multimodal encoders.
Taking consideration of scalability, previous methods~\citep{Wu2018MultimodalLearning,shi2019variational,NEURIPS2020_43bb733c,sutter2021generalized} proposed to approximate the multimodal encoders using the unimodal encoders instead of introducing the extra multimodal encoders, since it would otherwise require $2^{M - 1}$ encoders in total. There are two common choices to achieve this, the product of experts (PoE) and the mixture of experts (MoE). To align all encoder distributions simultaneously, PoE should therefore be avoided, as the product of experts is sharper than any of its experts. This makes it impossible to align different encoder distributions. 

Since our goal is to align all unimodal and multimodal encoders, we only need to align the unimodal encoder distributions and approximate the multimodal encoders via any abstract mean function~\citep{DBLP:journals/entropy/Nielsen20}. The multimodal encoder distributions are aligned automatically by aligning the unimodal encoder distributions alone as $q(\bm{z} | \bm{X}_K) = \frac {1}{K} \sum_{i=1}^K q(\bm{z} | \bm{x}_i) = q(\bm{z} | \bm{x}_1) = \dots = q(\bm{z} | \bm{x}_M)$, if we take the arithmetic mean (i.e., MoE) as the abstract mean function for example.
% they choose PoE because it decreases entropy However our analysis suggests avoiding PoE if cross generation is wanted. By using PoE, it forces every model only learns partial information 

As per the discussion above, we only need to align all unimodal encoder distributions and the decoder distribution during training. Afterwards, we can compute the multimodal posterior in the closed-form through the appropriate mean function during testing.

\begin{figure}[t]

\centering
\scalebox{0.6}{
\begin{tikzpicture}[node distance=5mm,
funcnode/.style={rectangle, draw, minimum height=0.75cm, minimum width=2.5cm},
tagnode/.style={minimum height=0.75cm, minimum width=2.5cm}]
    \node[funcnode] (Gz1) {$\hat{\bm{z}}_1 = G_{\bm{z}}(\bm{x}_1)$};
    \node[funcnode, below=0.5 of Gz1] (Gz2) {$\hat{\bm{z}}_2 = G_{\bm{z}}(\bm{x}_2)$};
    \node[funcnode, below=1.25 of Gz2] (GzM) {$\hat{\bm{z}}_M = G_{\bm{z}}(\bm{x}_M)$};
    \node at ($(Gz2)!.5!(GzM)$) {\vdots};
    
    \node[tagnode, right=1.25 of Gz1] (tupleGz1) {($\bm{X}, \hat{\bm{z}}_1) \sim q_1(\bm{X}, \bm{z})$};
    \node[tagnode, below=0.5 of tupleGz1] (tupleGz2) {($\bm{X}, \hat{\bm{z}}_2) \sim q_2(\bm{X}, \bm{z})$};
    \node[tagnode, below=1.25 of tupleGz2] (tupleGzM) {($\bm{X}, \hat{\bm{z}}_M) \sim q_M(\bm{X}, \bm{z})$};
    \node at ($(tupleGz2)!.5!(tupleGzM)$) {\vdots};
    
    \node[tagnode, left=of Gz1] (X) {$\bm{X} \sim q(\bm{X})$};
    
    \edge {X} {Gz1};
    \edge {X} {Gz2};
    \edge {X} {GzM};
    \edge {Gz1} {tupleGz1};
    \edge {Gz2} {tupleGz2};
    \edge {GzM} {tupleGzM};
    
    \node[funcnode, below=of GzM] (Gx1) {$\Tilde{\bm{x}}_1 = G_{\bm{x}_1}(\bm{z})$};
    \node[funcnode, below=0.5 of Gx1] (Gx2) {$\Tilde{\bm{x}}_2 = G_{\bm{x}_2}(\bm{z})$};
    \node[funcnode, below=1.25 of Gx2] (GxM) {$\Tilde{\bm{x}}_M = G_{\bm{x}_M}(\bm{z})$};
    \node at ($(Gx2)!.5!(GxM)$) {\vdots};

    \node[tagnode, below=0.5 of tupleGzM] (tupleGX) {($\Tilde{\bm{X}}, \bm{z}) \sim p(\bm{X}, \bm{z})$};
    
    \node[tagnode, left=of Gx1] (z) {$\bm{z} \sim p(\bm{z})$};

    \edge {z} {Gx1};
    \edge {z} {Gx2};
    \edge {z} {GxM};
    \edge {Gx1} {tupleGX};
    \edge {Gx2} {tupleGX};
    \edge {GxM} {tupleGX};
\end{tikzpicture}
}
\caption{The encoder distributions and the decoder distribution to be aligned. The discriminator $D(\bm{X}, \bm{z})$ tries to distinguish among tuples $(\bm{X}, \hat{\bm{z}}_i)$ and $(\Tilde{\bm{X}}, \bm{z})$.}
\label{fig:diagram}
\end{figure}
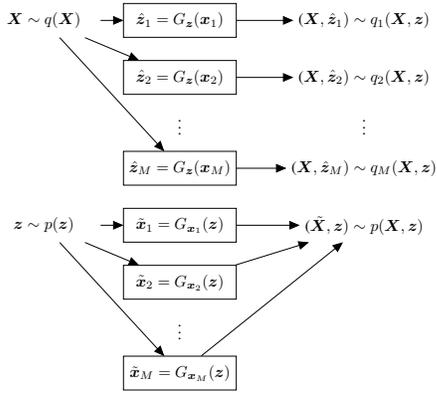

%We thus propose the following objective:
%\begin{align}
%    \min_{q_1, \dots, q_M, p} D_{\text{JSD}}(q_1(\bm{X}, \bm{z}), \dots, q_M(\bm{X}, \bm{z}), p(\bm{X}, \bm{z}))
%\end{align}
%where $D_{\text{JSD}}$ is the Jensen–Shannon divergence for multiple distributions.
To achieve this goal, we resort to adversarial learning, resulting in \textbf{M}ulti\textbf{M}odal \textbf{A}dversarially \textbf{L}earned \textbf{I}nference:
%To achieve this, we can set up an adversarial game with $M + 1$ distributions:
\begin{align}
\begin{split}
    \min_G \max_D V(D, G) = 
     \mathbb{E}_{p(\bm{z})} [\log D(G_{\bm{X}}(\bm{z}), \bm{z})[M + 1]] \\
 + \sum_{i=1}^M \mathbb{E}_{q(\bm{X})}[\log D(\bm{X}, G_{\bm{z}}(\bm{x}_i))[i]]
\end{split}
\end{align}
where $D$ is the discriminator that takes both $\bm{X}$ and $\bm{z}$ as input and outputs $M + 1$ probabilities. $G$ represents the corresponding unimodal encoders and decoders with a slight abuse of notation. Unlike ordinary GANs that only align two distributions, our proposed approach aligns $M+1$ distributions. It can be shown that this objective is equivalent to minimizing the Jensen–Shannon divergence of all distributions~\citep{pmlr-v80-pu18a}.

% Our analysis suggests, if we want translation among different modalities through the shared code, then the distribution of the shared representation must be aligned. Unlike previous method, which suggests using PoE to Boost information.

\subsection{Discriminator Factorization}
\label{sec:dis_factor}

Given the model specification discussed in Section~\ref{sec:mmali}, we show that a specific form of the discriminator can be constructed instead of a monolithic one.
% we propose a specific form of the discriminator
% In addition to fully observed multimodal data, one can usually gather more unimodal data in practice. To incorporate the unimodal data, we exploit the density ratio trick to factorize the joint discriminator.

It can be shown that the optimal discriminator $D^\ast$ given the fixed encoders and decoders is:
\begin{align}
  D^\ast(\bm{X}, \bm{z})[i] =
      \begin{cases}
          \frac{q_i(\bm{X}, \bm{z})}{p(\bm{X}, \bm{z}) + \sum_j^M q_j(\bm{X}, \bm{z})}, & 1 \le i \le M \\
          \frac{p(\bm{X}, \bm{z})}{p(\bm{X}, \bm{z}) + \sum_j^M q_j(\bm{X}, \bm{z})}, & i = M + 1
      \end{cases}
\end{align}
The proof can be found in~\citet{pmlr-v80-pu18a}.
This optimal joint discriminator can be expressed with the probability density ratios $r_i(\bm{X}, \bm{z}) = \frac{q_i(\bm{X}, \bm{z})}{p(\bm{X}, \bm{z})}$:
\begin{align}
  D^\ast(\bm{X}, \bm{z})[i] =
      \begin{cases}
          \frac{r_i(\bm{X}, \bm{z})}{1 + \sum_j^M r_j(\bm{X}, \bm{z})}, & 1 \le i \le M \\
          \frac{1}{1 + \sum_j^M r_j(\bm{X}, \bm{z})}, & i = M + 1
      \end{cases}
\end{align}
According to the model specification discussed before, for each $r_i(\bm{X}, \bm{z})$, we have:
\begin{align}
\begin{split}
    r_i(\bm{X}, \bm{z}) &=  \frac{q_i(\bm{X}, \bm{z})}{p(\bm{X}, \bm{z})} \\
    &= \frac{q(\bm{x}_1, \dots, \bm{x}_M)q(\bm{z} | \bm{x}_i)}{p(\bm{z}) \prod_j^M p(\bm{x}_j | \bm{z})} \\
    &= \frac{q(\bm{x}_1, \dots, \bm{x}_M)}{\prod_k^M q(\bm{x}_k)} \frac{q(\bm{z} | \bm{x}_i)}{p(\bm{z})} \prod_j^M \frac{q(\bm{x}_j)}{p(\bm{x}_j | \bm{z})} \\
    &= \frac{q(\bm{X})}{\prod_k^M q(\bm{x}_k)} \frac{q(\bm{x}_i)q(\bm{z} | \bm{x}_i)}{q(\bm{x}_i)p(\bm{z})} \prod_j^M \frac{q(\bm{x}_j)p(\bm{z})}{p(\bm{x}_j | \bm{z})p(\bm{z})} \\
    &= \frac{q(\bm{X})}{\prod_k^M q(\bm{x}_k)} \frac{q(\bm{x}_i, \bm{z})}{q(\bm{x}_i) p(\bm{z})} \prod_j^M \frac{q(\bm{x}_j)p(\bm{z})}{p(\bm{x}_j, \bm{z})}
\end{split}
\end{align}
%Taking advantages of the interpretation of probability density ratio estimation of GAN, 
Now we factorize the joint discriminator into the smaller discriminators that can be trained separately on the unimodal data. For each modality, we train two discriminators. To estimate $\frac{q(\bm{x}_i, \bm{z})}{q(\bm{x}_i) p(\bm{z})}$, we take samples from $q(\bm{x}_i, \bm{z})$ as positive and samples from $q(\bm{x}_i) p(\bm{z})$ as negative. To estimate $\frac{q(\bm{x}_i, \bm{z})}{p(\bm{x}_i, \bm{z})}$, we take samples from $q(\bm{x}_i, \bm{z})$ as positive and samples from $p(\bm{x}_i, \bm{z})$ as negative. $\frac{q(\bm{x}_i)p(\bm{z})}{p(\bm{x}_i, \bm{z})}$ is then computed by $\frac{q(\bm{x}_i, \bm{z})}{p(\bm{x}_i, \bm{z})} / \frac{q(\bm{x}_i, \bm{z})}{q(\bm{x}_i) p(\bm{z})}$. As for $\frac{q(\bm{x}_1, \dots, \bm{x}_M)}{\prod_k^M q(\bm{x}_k)}$, we train the separate discriminator by taking samples from $q(\bm{X})$ as positive and samples from the product of marginals $\prod_k^M q(\bm{x}_k)$ as negative. The advantages of such factorization lies in twofold. Firstly, the discriminator for $\frac{q(\bm{X})}{\prod_k^M q(\bm{x}_k)}$ can utilize multimodal data efficiently by learning from unpaired samples as well. Secondly, we can update encoders and decoders on unimodal data, since we now have probability density ratios $\frac{q(\bm{x}_i, \bm{z})}{p(\bm{x}_i, \bm{z})}$.

A special case for $\frac{q(\bm{X})}{\prod_k^M q(\bm{x}_k)}$ in the two-modality setting is related to mutual information maximization. Note that each estimated probability density ratio $r_i(\bm{X}, \bm{z}) \propto \frac{q(\bm{x}_1, \bm{x}_2)}{q(\bm{x}_1) q(\bm{x}_2)}$.
By updating the decoder distribution to minimize $\mathbb{E}_{p(\bm{X}, \bm{z})} \log D^\ast(\bm{X}, \bm{z})[M+1] = -\mathbb{E}_{p(\bm{X}, \bm{z})} \log (1 + \sum_i^M r_i(\bm{X}, \bm{z}))$, the decoders are updated to maximize the mutual information of two modalities implicitly.
% Benefiting from such factorization, we can combine multimodal and unimodal training under the same framework. Furthermore, the factorized discriminator $\frac{q(\bm{X})}{\prod_j q(\bm{x}_j)}$ embeds a contrastive learning mechanism in our framework.
% As we align all distributions, the probability density ratios $r_i(\bm{X}, \bm{z})$ are pushed to $1$, which in turn forces our model maximization the mutual information implicitly.
% decoder distribution is updated to minimize $\mathbb{E} \log D^\ast = -\mathbb{E} \log (1 + \sum r_i) $ which in turn maximize the mutual information of generated samples.
% Inspecting the term, such loss is related to the mutual information maximization. Minimizing the divergence of all distributions will push $\frac{q_i}{p}$ to $1$, the decoders will maximize $\frac{q(x_1, x_2)}{q(x_1), q(x_2)}$. which means decoders maximize the mutual information implicitly.

\subsection{Latent Space Factorization}
\label{sec:latent_factor}

% The model discussed in the previous section has assumed stochastic mappings for each conditional $p(\bm{x}_i | \bm{z})$ and $q(\bm{x}_i | \bm{z})$.
As discussed in the previous section, we force different unimodal encoders to have the same conditional distribution $q(\bm{X})q(\bm{z} | \bm{x}_i) = q(\bm{X})q(\bm{z} | \bm{x}_j)$, which may make the modality-specific information lost. Our decoders are the deterministic mappings parameterized by neural networks. Without extra information, the model will struggle to balance between information preserving and distribution matching. We therefore introduce the modality-specific codes to alleviate this problem.

By introducing the modality-specific style codes $\bm{S} = \{\bm{s}_i\}_{i=1}^M$ and the shared content code $\bm{c}$, with the assumption that modality-specific style codes $\bm{S}$ and the shared content code $\bm{c}$ are conditionally independent given the observed data $\bm{X}$. The unimodal encoder distributions can derived as below:
\begin{align}
\begin{split}
        q_i(\bm{X}, \bm{S}, \bm{c}) &=
        q(\bm{X})q(\bm{c} | \bm{x}_i)\prod_j^M q(\bm{s}_j | \bm{x}_j)
\end{split}
\end{align}
and the decoder distribution is as follows:
\begin{align}
\begin{split}
        p(\bm{X}, \bm{S}, \bm{c}) &=
        p(\bm{c}) \prod_i^M p(\bm{s}_i) p(\bm{x}_i | \bm{s}_i, \bm{c})
\end{split}
\end{align}

Following the new model specification, the similar result can be obtained for $\frac{q_i(\bm{X}, \bm{S}, \bm{c})}{p(\bm{X}, \bm{S}, \bm{c})}$:
\begin{align}
\begin{split}
    \frac{q_i(\bm{X}, \bm{S}, \bm{c})}{p(\bm{X}, \bm{S}, \bm{c})}
    % &= \\
    &= \frac{q(\bm{x}_1, \dots, \bm{x}_M)q(\bm{c} | \bm{x}_i)\prod_j^M q(\bm{s}_j | \bm{x}_j)}{p(\bm{c}) \prod_k^M p(\bm{s}_k) p(\bm{x}_k | \bm{s}_k, \bm{c})} \\
    &= \frac{q(\bm{X})}{\prod_k^M q(\bm{x}_k)} \frac{q(\bm{c} | \bm{x}_i)}{p(\bm{c})} \prod_j^M \frac{q(\bm{x}_j) q(\bm{s}_j | \bm{x}_j)}{p(\bm{s}_j)p(\bm{x}_j|\bm{s}_j, \bm{c})} \\
    % &= \frac{q(\bm{X})}{\prod_k^M q(\bm{x}_k)} \frac{q(\bm{c} | \bm{x}_i) q(\bm{s_i} | \bm{x}_i) q(\bm{x}_i)}{q(\bm{s_i} | \bm{x}_i)q(\bm{x}_i)p(\bm{c})} \prod_j^M \frac{q(\bm{x}_j) q(\bm{s}_j | \bm{x}_j)}{p(\bm{s}_j)p(\bm{x}_j|\bm{s}_j, \bm{c})} \\
    &= \frac{q(\bm{X})}{\prod_k^M q(\bm{x}_k)} 
    \frac{q(\bm{x}_i, \bm{s}_i, \bm{c})}{q(\bm{x}_i, \bm{s}_i) p(\bm{c})} \prod_j^M \frac{q(\bm{x}_j, \bm{s}_j)p(\bm{c})}{p(\bm{x}_j, \bm{s}_j, \bm{c})}
\end{split}
\end{align}
% The more detailed derivation is given in the appendix. 
Now we can train the factorized discriminators as discussed above.

\section{Experiments}
\label{sec:exps}

To examine the effectiveness of our proposed approach, we conduct experiments on three different datasets. Across all the experiments, we use Adam optimizer~\citep{DBLP:journals/corr/KingmaB14} with the learning rate $0.0002$, in which all models are trained for $250,000$ iterations with the batch size $64$. We also employ an exponential moving average~\citep{DBLP:conf/iclr/YaziciFWYPC19} of the weights for both encoders and decoders with a decay rate $0.9999$ and start at the $50,000$-th iteration. In all models, the standard Gaussian $\mathcal{N}(\bm{0},\bm{I})$ is chosen as the prior, and the isotropic Gaussian $\mathcal{N}(\bm{\mu}, \bm{\sigma}^2\bm{I})$ is chosen as the posterior. We make use of the non-saturating loss described in \citet{Dandi_Bharadhwaj_Kumar_Rai_2021} to update the encoders and decoders.

%We evalute the proposed model following the similar setting of \citet{shi2019variational} and \citet{shi2021relating}. Specifically, following metrics are adopted to evalute our model:

\subsection{MultiMNIST}
In this experiment, we compare the model trained with the factorized discriminator and the joint discriminator to investigate the effectiveness of discriminator factorization.

\begin{figure}[h]
    \centering
    \includegraphics[scale=0.5]{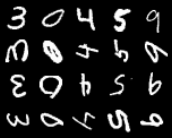}
    \caption{5 samples of 4-modality MultiMNIST dataset.}
\end{figure}

\paragraph{Setup} MultiMNIST dataset consists of multiple modalities, where every modality is a rotated version of MNIST~\citep{6296535}. Every instance in one modality is associated with 30 random instances of the same digit class from other modalities. We perform experiments up to four modalities, in which each modality is rotated counterclockwise $0, 90, 180, 270$ degrees, respectively. The performance is assessed by training a linear classifier on top of the frozen encoder of the original MNIST modality (rotated 0 degrees). In our empirical study, the same architecture of the encoders and the decoders is used for each modality across every experiments.

\begin{table}[h]
    \centering
    \begin{tabular}{r c c c}
        \toprule
        \multirow{2}{*}{\textbf{Methods}} & \multicolumn{3}{c}{\textbf{Modalities ($\bm{\%}$)}} \\
        \cline{2-4} \\
        & 2 & 3 & 4 \\
        \midrule
        Joint & 85.96 & 85.44 & 85.27 \\
        Factorized & 97.02 & 96.94 & 96.75 \\
        \bottomrule
    \end{tabular}
    \caption{The classification accuracy of the representations of encoders trained with different discriminators.}
    \label{tab:multimnist}    
\end{table}

\begin{figure*}[h]
\centering
    \includegraphics[scale=0.6]{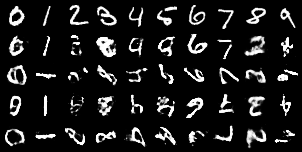}
    % \caption{(a)}
    \includegraphics[scale=0.6]{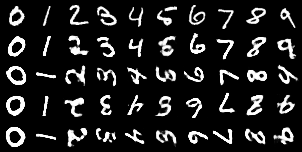}
    % \caption{(b)}
    % \includegraphics[scale=0.35]{joint_random_generation.png}
    % \includegraphics[scale=0.35]{factor_random_generation.png}
    \caption{Left: Joint 
    Right: Factorized. From top to bottom: original, reconstruction, translation to 90, translation to 180, translation to 270.}
    \label{fig:multiminist}
\end{figure*}

Table~\ref{tab:multimnist} shows that the performance of the representations of encoders trained with factorized discriminators is improved significantly. This experiment also demonstrates the generalizability on more than two modalities for the proposed method. Additionally, the qualitative results are shown in Figure~\ref{fig:multiminist}.

\subsection{MNIST-SVHN}

\begin{table*}[t]
% \begin{minipage}{\linewidth}
\begin{center}
    \scalebox{0.8}{
        \begin{tabular}{ @{}crccccccc@{} }
        \toprule
        \multirow{2}{*}{\textbf{Data}} &
        \multirow{2}{*}{\textbf{Methods}} & \multicolumn{2}{c}{\textbf{Latent ($\bm{\%}$)}} & \multirow{2}{*}{\textbf{Joint ($\bm{\%}$)}} & \multicolumn{2}{c}{\textbf{Cross Generation ($\bm{\%}$)}} & \multicolumn{2}{c}{\textbf{Synergy ($\bm{\%}$)}} \\
        \cline{3-4} \cline{6-7} \cline{8-9} && \textbf{M} & \textbf{S} && \textbf{S} $\rightarrow$ \textbf{M} & \textbf{M} $\rightarrow$ \textbf{S} & \textbf{M} & \textbf{S} \\
        \midrule
        \multirow{10}{*}{100\%}
        &JMVAE & 84.45 (\textit{\scriptsize$\pm$\scriptsize0.87}) & 57.98 (\textit{\scriptsize$\pm$\scriptsize1.27}) & 42.18 (\textit{\scriptsize$\pm$\scriptsize1.50}) & 49.63 (\textit{\scriptsize$\pm$\scriptsize1.78}) & 54.98 (\textit{\scriptsize$\pm$\scriptsize3.02}) & 85.77 (\textit{\scriptsize$\pm$\scriptsize0.66}) & 68.15 (\textit{\scriptsize$\pm$\scriptsize1.38}) \\
        &cI-JMVAE & 84.58 (\textit{\scriptsize$\pm$\scriptsize1.49}) & 64.42 (\textit{\scriptsize$\pm$\scriptsize1.42}) & 48.95 (\textit{\scriptsize$\pm$\scriptsize2.31}) & 58.16 (\textit{\scriptsize$\pm$\scriptsize1.83}) & 70.61 (\textit{\scriptsize$\pm$\scriptsize3.13}) & 93.45 (\textit{\scriptsize$\pm$\scriptsize0.52}) & 84.00 (\textit{\scriptsize$\pm$\scriptsize0.97}) \\
        &cC-JMVAE & 83.67 (\textit{\scriptsize$\pm$\scriptsize3.48}) & 66.64 (\textit{\scriptsize$\pm$\scriptsize2.92}) & 47.27 (\textit{\scriptsize$\pm$\scriptsize4.52}) & 59.73 (\textit{\scriptsize$\pm$\scriptsize3.85}) & 69.49 (\textit{\scriptsize$\pm$\scriptsize2.19}) & 91.21 (\textit{\scriptsize$\pm$\scriptsize6.59}) & 84.07 (\textit{\scriptsize$\pm$\scriptsize3.19}) \\
        &MVAE & 91.65 (\textit{\scriptsize$\pm$\scriptsize0.17}) & 64.12 (\textit{\scriptsize$\pm$\scriptsize4.58}) & 9.42 (\textit{\scriptsize$\pm$\scriptsize7.82}) & 10.98 (\textit{\scriptsize$\pm$\scriptsize0.56}) & 21.88 (\textit{\scriptsize$\pm$\scriptsize2.21}) & 64.60 (\textit{\scriptsize$\pm$\scriptsize9.25}) & 52.91 (\textit{\scriptsize$\pm$\scriptsize8.11}) \\
        &cI-MVAE & 96.97 (\textit{\scriptsize$\pm$\scriptsize0.84}) & 75.94 (\textit{\scriptsize$\pm$\scriptsize6.20}) & 15.23 (\textit{\scriptsize$\pm$\scriptsize10.46}) & 10.85 (\textit{\scriptsize$\pm$\scriptsize1.17}) & 27.70 (\textit{\scriptsize$\pm$\scriptsize2.09}) & {85.07 (\textit{\scriptsize$\pm$\scriptsize7.73})} & {75.67 (\textit{\scriptsize$\pm$\scriptsize4.13})} \\
        &cC-MVAE & \textbf{97.42 (\textit{\scriptsize$\pm$\scriptsize0.40})} & 81.07 (\textit{\scriptsize$\pm$\scriptsize2.03}) & 8.85 (\textit{\scriptsize$\pm$\scriptsize3.86}) & 12.83 (\textit{\scriptsize$\pm$\scriptsize2.25}) & 30.03 (\textit{\scriptsize$\pm$\scriptsize2.46}) & {75.25 (\textit{\scriptsize$\pm$\scriptsize5.31})} & {69.42 (\textit{\scriptsize$\pm$\scriptsize3.94})}\\
        & MMVAE & 92.48 (\textit{\scriptsize$\pm$\scriptsize0.37}) & 79.03 (\textit{\scriptsize$\pm$\scriptsize1.17}) & 42.32 (\textit{\scriptsize$\pm$\scriptsize2.97}) & 70.77 (\textit{\scriptsize$\pm$\scriptsize0.35}) & 85.50 (\textit{\scriptsize$\pm$\scriptsize1.05}) & --- & --- \\
        &cI-MMVAE & 93.97 (\textit{\scriptsize$\pm$\scriptsize0.36}) & 81.87 (\textit{\scriptsize$\pm$\scriptsize0.52}) & {43.94 (\textit{\scriptsize$\pm$\scriptsize0.96})} & 79.66 (\textit{\scriptsize$\pm$\scriptsize0.59}) & 92.67 (\textit{\scriptsize$\pm$\scriptsize0.29}) & --- & --- \\
        &cC-MMVAE & {93.10 (\textit{\scriptsize$\pm$\scriptsize0.17})} & {80.88 (\textit{\scriptsize$\pm$\scriptsize0.80})} & 45.46 (\textit{\scriptsize$\pm$\scriptsize0.78}) & {79.34 (\textit{\scriptsize$\pm$\scriptsize0.54})} & {92.35 (\textit{\scriptsize$\pm$\scriptsize0.46})} & --- & --- \\
        & MMJSD (MS) & {95.76 (\textit{\scriptsize$\pm$\scriptsize0.20})} & {78.12 (\textit{\scriptsize$\pm$\scriptsize0.83})} & {25.33 (\textit{\scriptsize$\pm$\scriptsize1.30})} & {45.81 (\textit{\scriptsize$\pm$\scriptsize4.09})} & {86.33 (\textit{\scriptsize$\pm$\scriptsize0.72})} & {89.66 (\textit{\scriptsize$\pm$\scriptsize0.18})} & {82.62 (\textit{\scriptsize$\pm$\scriptsize0.49})} \\
        & \textbf{MMALI (Ours)} & 96.82 (\textit{\scriptsize$\pm$\scriptsize0.59}) & \textbf{87.20 (\textit{\scriptsize$\pm$\scriptsize0.75})} & \textbf{79.31 (\textit{\scriptsize$\pm$\scriptsize0.01})} & \textbf{83.80 (\textit{\scriptsize$\pm$\scriptsize0.01})} & \textbf{95.22 (\textit{\scriptsize$\pm$\scriptsize0.02})} & \textbf{96.43 (\textit{\scriptsize$\pm$\scriptsize0.01})} & \textbf{96.69 (\textit{\scriptsize$\pm$\scriptsize0.01})} \\
        
        \midrule
        \multirow{10}{*}{20\%}
        &JMVAE & 77.53 (\textit{\scriptsize$\pm$\scriptsize0.13}) & 52.55 (\textit{\scriptsize$\pm$\scriptsize2.18}) & 26.37 (\textit{\scriptsize$\pm$\scriptsize0.54}) & 42.58 (\textit{\scriptsize$\pm$\scriptsize5.32}) & 41.44 (\textit{\scriptsize$\pm$\scriptsize2.26}) & 85.07 (\textit{\scriptsize$\pm$\scriptsize9.74}) & 51.95 (\textit{\scriptsize$\pm$\scriptsize2.28}) \\
        &cI-JMVAE & 77.57 (\textit{\scriptsize$\pm$\scriptsize4.02}) & 57.91 (\textit{\scriptsize$\pm$\scriptsize1.28}) & 32.58 (\textit{\scriptsize$\pm$\scriptsize5.89}) & 51.85 (\textit{\scriptsize$\pm$\scriptsize1.27}) & 47.92 (\textit{\scriptsize$\pm$\scriptsize10.32}) & \textbf{92.54 (\textit{\scriptsize$\pm$\scriptsize1.13})} & 67.01 (\textit{\scriptsize$\pm$\scriptsize8.72}) \\
        &cC-JMVAE & 81.11 (\textit{\scriptsize$\pm$\scriptsize2.76}) & 57.85 (\textit{\scriptsize$\pm$\scriptsize2.23}) & 34.00 (\textit{\scriptsize$\pm$\scriptsize7.18}) & 50.73 (\textit{\scriptsize$\pm$\scriptsize0.45}) & 56.89 (\textit{\scriptsize$\pm$\scriptsize6.18}) & 88.36 (\textit{\scriptsize$\pm$\scriptsize4.38}) & 68.49 (\textit{\scriptsize$\pm$\scriptsize8.82}) \\
        &MVAE & 90.29 (\textit{\scriptsize$\pm$\scriptsize0.57}) & 33.44 (\textit{\scriptsize$\pm$\scriptsize0.26}) & 10.88 (\textit{\scriptsize$\pm$\scriptsize9.15}) & 8.72 (\textit{\scriptsize$\pm$\scriptsize0.92}) & 12.12 (\textit{\scriptsize$\pm$\scriptsize3.38}) & 42.10 (\textit{\scriptsize$\pm$\scriptsize5.22}) & 44.95 (\textit{\scriptsize$\pm$\scriptsize5.92}) \\
        &cI-MVAE & 93.72 (\textit{\scriptsize$\pm$\scriptsize1.09}) & 56.74 (\textit{\scriptsize$\pm$\scriptsize7.97}) & 12.79 (\textit{\scriptsize$\pm$\scriptsize6.82}) & 14.18 (\textit{\scriptsize$\pm$\scriptsize2.19}) & 20.23 (\textit{\scriptsize$\pm$\scriptsize4.55}) & {75.36 (\textit{\scriptsize$\pm$\scriptsize5.05})} &  {64.81 (\textit{\scriptsize$\pm$\scriptsize4.81})} \\
        &cC-MVAE & {92.74 (\textit{\scriptsize$\pm$\scriptsize2.97})} & 52.99 (\textit{\scriptsize$\pm$\scriptsize8.35}) & 17.95 (\textit{\scriptsize$\pm$\scriptsize12.52}) & 14.70 (\textit{\scriptsize$\pm$\scriptsize1.65}) & 24.90 (\textit{\scriptsize$\pm$\scriptsize5.77}) & {56.86 (\textit{\scriptsize$\pm$\scriptsize18.84})} &  {54.28 (\textit{\scriptsize$\pm$\scriptsize9.86})} \\ 
        &MMVAE & 88.54 (\textit{\scriptsize$\pm$\scriptsize0.37}) & 68.90 (\textit{\scriptsize$\pm$\scriptsize1.79}) & 37.71 (\textit{\scriptsize$\pm$\scriptsize0.60}) & 59.52 (\textit{\scriptsize$\pm$\scriptsize0.28}) & 76.33 (\textit{\scriptsize$\pm$\scriptsize2.23}) & --- & --- \\
        &cI-MMVAE & 91.64 (\textit{\scriptsize$\pm$\scriptsize0.06}) & 73.02 (\textit{\scriptsize$\pm$\scriptsize0.80}) & 42.74 (\textit{\scriptsize$\pm$\scriptsize0.36}) & 69.51 (\textit{\scriptsize$\pm$\scriptsize1.18}) & \textbf{86.75 (\textit{\scriptsize$\pm$\scriptsize0.28})} & --- & --- \\
        &{cC-MMVAE} & 92.10 (\textit{\scriptsize$\pm$\scriptsize0.19}) & {71.29 (\textit{\scriptsize$\pm$\scriptsize1.05})} & {40.77 (\textit{\scriptsize$\pm$\scriptsize0.93})} & {68.43 (\textit{\scriptsize$\pm$\scriptsize0.90})} & {86.24 (\textit{\scriptsize$\pm$\scriptsize0.89})} & --- & --- \\ 
        & MMJSD (MS) & {92.78 (\textit{\scriptsize$\pm$\scriptsize0.07})} & {61.17 (\textit{\scriptsize$\pm$\scriptsize1.62})} & {19.29 (\textit{\scriptsize$\pm$\scriptsize1.43})} & {29.17 (\textit{\scriptsize$\pm$\scriptsize1.17})} & {68.05 (\textit{\scriptsize$\pm$\scriptsize0.75})} & {87.97 (\textit{\scriptsize$\pm$\scriptsize0.51})} & {64.85 (\textit{\scriptsize$\pm$\scriptsize0.89})} \\
        & \textbf{MMALI (Ours)} & \textbf{94.03 (\textit{\scriptsize$\pm$\scriptsize0.64})} & \textbf{80.16 (\textit{\scriptsize$\pm$\scriptsize0.74})} & \textbf{75.88 (\textit{\scriptsize$\pm$\scriptsize0.01})} & \textbf{72.83 (\textit{\scriptsize$\pm$\scriptsize0.02})} & 86.19 (\textit{\scriptsize$\pm$\scriptsize0.03}) & 84.21 (\textit{\scriptsize$\pm$\scriptsize0.02})  & \textbf{85.24 (\textit{\scriptsize$\pm$\scriptsize0.01})} \\        
        \bottomrule
    \end{tabular}}
\end{center}
% \end{minipage}

\caption{Baselines JMVAE, MVAE, MMVAE, their contrastive variations, MMJSD with modality-specific code and our method (I stands for IWAE~\citep{DBLP:journals/corr/BurdaGS15} and C stands for CUBO~\citep{NIPS2017_35464c84}). M stands for MNIST, and S represents SVHN.}
\label{tab:mnist_svhn_results}
\end{table*}

We compare the proposed method with the-state-of-the-art methods on MNIST-SVHN.
\paragraph{Setup} MNIST-SVHN dataset contains two modalities, MNIST and SVHN~\citep{37648}, where each instance of a digit class is paired with 30 random instances from the other dataset with the same digit class. We use the same architecture of encoders and decoders used by \citet{shi2021relating}, where MLPs are used for MNIST and CNNs are used for SVHN. Furthermore, we use the same 20-d for the latent space while splitting it into the 10-d content code and 10-d style code. As suggested in \citet{shi2019variational,shi2021relating}, the following metrics are used to evaluate the performance of the models.
\begin{itemize}
    \item \textbf{Latent accuracy} Latent accuracy is computed by the classification accuracy of a linear classifier trained on the learned representation of encoders.
    \item \textbf{Joint coherence accuracy} Joint coherence accuracy is computed by how often multimodal generations are classified to have the same label. We pre-train the classifiers of each modality and apply them on the multimodal generations $\bm{x}_1, \bm{x}_2 \sim p(\bm{z}) p(\bm{x}_1, \bm{x}_2 | \bm{z})$ to check how often $\bm{x}_1$ and $\bm{x}_2$ have the same label.
    \item \textbf{Cross coherence accuracy} Cross coherence accuracy is computed similarly to joint coherence accuracy. We apply the pre-trained classifiers on the cross generated samples $\bm{x}_1, \hat{\bm{x}}_2 \sim q(\bm{x}_1) q(\bm{z} | \bm{x}_1) p(\hat{\bm{x}}_2 | \bm{z})$ and $\hat{\bm{x}}_1, \bm{x}_2 \sim q(\bm{x}_2) q(\bm{z} | \bm{x}_2) p(\hat{\bm{x}}_1 | \bm{z})$ and compute how often the label remains the same.
    \item \textbf{Synergy coherence accuracy} Synergy coherence accuracy is computed by how often the joint reconstructions have the same label as the original input. The pre-trained classifiers are now applied on $\hat{\bm{x}}_1, \hat{\bm{x}}_2 \sim q(\bm{x}_1, \bm{x}_2) q(\bm{z} | \bm{x}_1, \bm{x}_2) p(\hat{\bm{x}}_1, \hat{\bm{x}}_2 | \bm{z})$
\end{itemize}

The experimental results are shown in Table~\ref{tab:mnist_svhn_results}. Moreover, we report the synergy coherence accuracy for our model by computing the joint posterior of Gaussian experts in the closed-form via geometric mean. It can be seen that our proposed model outperforms others in almost every metrics, especially on the joint coherence accuracy.

% We notice that the joint coherence is higher by a large margin. We think this is the result of the evaluation method used. as we use a classifier to assess the coherence. Since during training the decoders try to fool the discriminators, so this method may favor  adversarially-trained models.

\begin{figure*}[t]
    \centering
    \scalebox{0.8}{
        \begin{tabular}{cc}
            \includegraphics[scale=0.6]{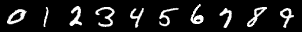} & \includegraphics[scale=0.7]{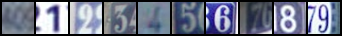} \\
            \includegraphics[scale=0.6]{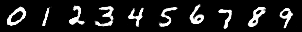} & \includegraphics[scale=0.7]{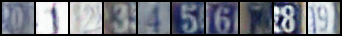} \\
            \includegraphics[scale=0.6]{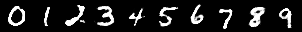} & \includegraphics[scale=0.7]{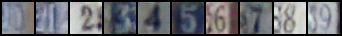} \\
            \includegraphics[scale=0.6]{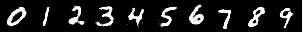} & \includegraphics[scale=0.7]{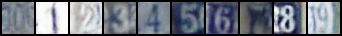} \\
            \includegraphics[scale=0.6]{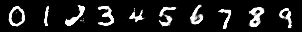} & \includegraphics[scale=0.7]{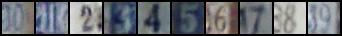} \\
        \end{tabular}
    }
    \caption{Latent space factorization. From top to bottom: original, reconstruction, reconstruction with random style, cross translation with original style, cross translation with random style.}
    \label{fig:mnist_svhn_rec}
\end{figure*}

\paragraph{Latent space factorization} We evaluate the effectiveness of latent space factorization. To this end, we train the linear classifiers on the content code, the style code and the whole latent code, respectively. Table~\ref{tab:content_style_code} shows that the classifiers trained on the whole latent code only outperform the classifiers trained on the content code marginally while performing significantly better than the classifiers trained on the style code. Figure~\ref{fig:mnist_svhn_rec} depicts the visual effect of latent space factorization. It can be seen that our proposed approach is able to effectively factorize the content from the style.

\begin{table}[h]
    \centering
    \begin{tabular}{@{}r c c c c@{}}
        % \multirow{2}{*}{\textbf{Model}} & \multicolumn{3}{c}{\textbf{Code (\%)}} \\
        % \cline{2-4} \\
        & \textbf{Content} & \textbf{Style} & \textbf{Both} \\
        \midrule
        MNIST & 96.61 &  23.56 &  96.88 \\
        SVHN &  86.40 &  21.39 &  86.49 \\
    \end{tabular}
    \caption{Classification accuracy ($\bm{\%}$) with different latent codes.}
    \label{tab:content_style_code}

\end{table}

\subsection{CUB Image-Captions}

We then evaluate our method on Caltech-UCSD Birds (CUB) dataset~\citep{WelinderEtal2010}, where each image of birds is annotated with 10 different captions. Since generating discrete data is challenging for GANs, we circumvent this by modeling it in the continuous feature space. The image features are obtained from the pre-trained ResNet-101 model~\citep{7780459}, and the features of caption are extracted from a pre-trained word-level CNN-LSTM autoencoder~\citep{NIPS2016_eb86d510}. 

\begin{table*}[h]
    \centering
    \scalebox{0.8}{
        \begin{tabular}{r c c c c}
            & \textbf{Joint} & \textbf{Cross (I $\rightarrow$ C)} & \textbf{Cross (C $\rightarrow$ I)} & \textbf{Ground Truth} \\
            \midrule
    
            MVAE  & -0.095 &  0.011 & -0.013 &  0.273 \\
            MMVAE &  0.263 &  0.104 &  0.135 &  0.273 \\
            MMALI (ours) &  0.502 &  0.488 &  0.545 &  0.406
       \end{tabular}
    }
    \caption{Correlation score of Image (I)-Caption (C) pair for joint and cross generation.}
    \label{tab:cub}
\end{table*}

We adopt the evaluation method employed by \citet{DBLP:journals/corr/abs-1812-06417,shi2019variational} to compute the correlation score as the groundtruth. We perform Canonical Correlation Analysis (CCA) on the feature space of two modalities, which learns two projection $W_1 \in \mathbb{R}^{n_1 \times k}$ and $W_2 \in \mathbb{R}^{n_2 \times k}$ mapping the observations $x_1 \in \mathbb{R}^{n_1}$ and $x_2 \in \mathbb{R}^{n_2}$ to the same dimension $W_1^T x_1$ and $W_2^T x_2$. The correlation score of new observations $\tilde{x}_1, \tilde{x}_2$ is then computed by:
\begin{align}
    \operatorname{corr}(\tilde{x}_1, \tilde{x}_2) = \frac{\phi(\tilde{x}_1)^T \phi(
    \tilde{x}_2)}{||\phi(\tilde{x}_1)||_2 ||\phi(
    \tilde{x}_2)||_2}
\end{align}
where $\phi(\tilde{x}_i) = W_i^T \tilde{x}_i - \operatorname{avg}(W_i^T x_i) $
% The evaluation metrics for this experiment is similar to that of \cite{shi2019variational}, except we use different caption embeddings for the caption. The projection matrix is computed on the training set of CUB Image-Captions.

\paragraph{Quantitative results} We perform CCA using different features to calculate the correlation score. Although the ground truth of the test set is different from \citet{shi2019variational}, we include the results of MMVAE for reference. Note that the correlation score of our model is higher than the ground truth score while MMVAE and MVAE are lower than theirs' score.

\begin{figure}[h]
    \centering
    \scalebox{0.8}{
            \begin{tabular}{m{4cm} m{4cm}}
                % \textbf{Input} & \textbf{Generated} \\
                \includegraphics[width=4cm, height=3cm]{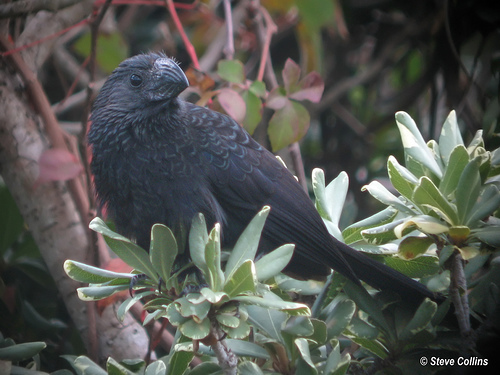} &
                {a small black bird with a long black beak and a black head} \\
                \includegraphics[width=4cm, height=3cm]{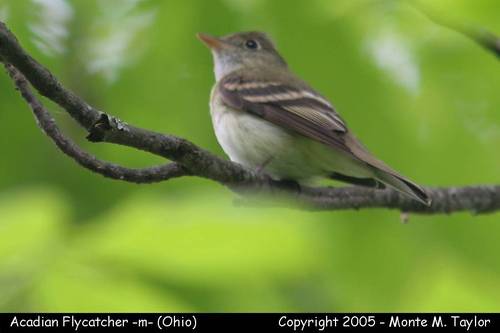} &
                {this particular bird has a white belly and breast and a small beak} \\
                \includegraphics[width=4cm, height=3cm]{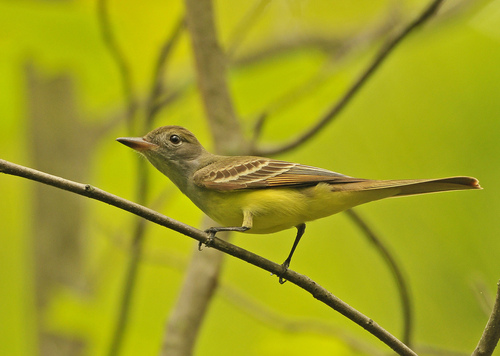} &
                {small yellow and brown bird with a long pointed beak and black and yellow feathers on its breast} \\
                % {this bird has wings that are brown and has a long neck} & \includegraphics[width=4cm, height=3cm]{Eared_Grebe_0027_34341.jpg} \\
                % {a black bird with a long wing span and a small beak} & \includegraphics[width=4cm, height=3cm]{Cape_Glossy_Starling_0096_129388.jpg} \\
                % {this bird is blue with black and has a very short beak} & \includegraphics[width=4cm, height=3cm]{Florida_Jay_0016_65051}
            \end{tabular}
        }
        \caption{From image to caption.}
        \label{fig:cub_cross1}
\end{figure}

\begin{figure}[h]
    \centering
    \scalebox{0.8}{
            \begin{tabular}{m{4cm} m{4cm}}
                % \textbf{Input} & \textbf{Generated} \\
                % \includegraphics[width=4cm, height=3cm]{Groove_Billed_Ani_0036_1604.jpg} &
                % {a small black bird with a long black beak and a black head} \\
                % \includegraphics[width=4cm, height=3cm]{Acadian_Flycatcher_0036_795577.jpg} &
                % {this particular bird has a white belly and breast and a small beak} \\
                % \includegraphics[width=4cm, height=3cm]{Great_Crested_Flycatcher_0124_29294.jpg} &
                % {small yellow and brown bird with a long pointed beak and black and yellow feathers on its breast} \\
                {this bird has wings that are brown and has a long neck} & \includegraphics[width=4cm, height=3cm]{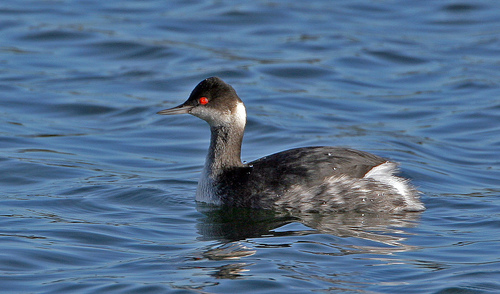} \\
                {a black bird with a long wing span and a small beak} & \includegraphics[width=4cm, height=3cm]{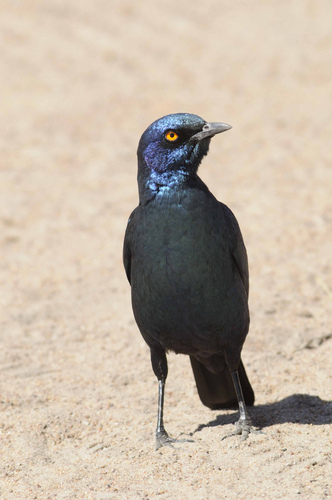} \\
                {this bird is blue with black and has a very short beak} & \includegraphics[width=4cm, height=3cm]{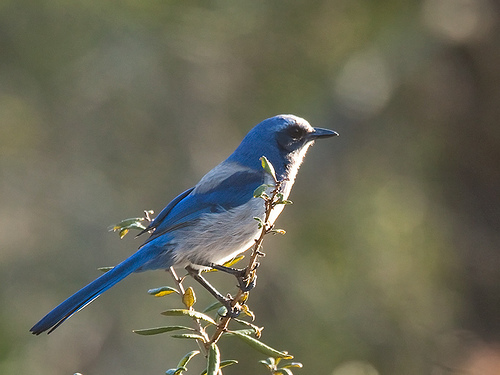}
            \end{tabular}
        }
        \caption{From caption to image.}
        \label{fig:cub_cross2}
\end{figure}

\paragraph{Qualitative results} To visualize the result, we project from the image feature space back to the pixel space by performing a nearest-neighbour lookup using Euclidean distance in the feature space on the training set. Moreover, we employ the pre-trained LSTM decoder to decode the generated caption features back to captions. Figure~\ref{fig:cub_cross1} and Figure~\ref{fig:cub_cross2} show the results of cross-modal generations in both directions.

\section{Related works}
\label{sec:related}

\paragraph{Multimodal VAEs} In the multimodal setting of VAEs, \citet{DBLP:conf/iclr/SuzukiNM17} introduced JVAE, where an extra multimodal encoder is trained to be aligned with other unimodal encoders. Similarly, \citet{vedantam2018generative} utilized an extra multimodal encoder, while their training procedure is split into two stages. Moreover, they  proposed to use PoE to handle the missing data during inference.
Since introducing the extra multimodal encoders explicitly is exponential in modalities, \citet{Wu2018MultimodalLearning} proposed MVAE to approximate the multimodal encoder through the product of experts. However, MVAE failed to translate from one modality to another. One of the motivations of PoE was to produce the sharper joint posterior, since it would increase the beliefs when more modalities are observed~\citep{vedantam2018generative, Kurle2019multi}. Our analysis suggests if we want to translate among different modalities, the latent code should be aligned.
Most recently, \citet{shi2019variational} proposed to replace PoE with MoE, which achieved the better cross generation performance. \citet{NEURIPS2020_43bb733c} enabled the efficient training by computing the joint posterior in closed-form, which eliminates inefficient sampling to approximate the joint posterior. Furthermore, \citet{shi2021relating} introduced a constrastive version of ELBO to improve data-efficiency. By introducing contrastive term, they make the model to utilize the unpaired data. Unfortunately, their objective is heuristically derived and empirically weighted to avoid the degenerate solutions, where the models only generates random noise. Instead, our model employs a discriminator to achieve constrastive learning.

% The motivation of PoE was to produce sharper joint posterior, but our analysis has suggested otherwise.
% and factorize the model instead
% the problem with $D(X)$ is that it does not see the fake samples during training

% The assumption of PoE is different. Different modalities provide complementary information to each other

% while previous works suggests using PoE increases belief, we show it otherwise

% the PoE was motivated by learn sharper posterior our analysis has however suggested otherwise.

\paragraph{GANs with specific form of discriminators} \citet{miyato2018cgans} proposed a specific form for the discriminator of cGAN. Instead of concatenating the conditional vector to the feature vectors, they constructed the discriminator in the form of the dot product of the conditional vector and the feature vector with certain
regularity assumption. They showed that a specific form of discriminator can improve the conditional image generation quality significantly. Our approach is inspired by FactorGAN~\citep{Stoller2020Training}, where they proposed to factorize the joint data distribution into a set of lower-dimensional distributions. Such factorization allows the model to be trained on incomplete observation. The factorization in~\citet{Stoller2020Training} is quite arbitrary while our factorization naturally follows the model specification.

\paragraph{Factorized latent space} Factorizing the latent space into modality-specific and shared codes is not a new idea. In image-to-image translation, MUNIT~\citep{10.1007/978-3-030-01219-9_11} introduced content and style codes, which aimed at achieving many-to-many translation. In the multimodal generative modeling, \citet{tsai2018learning} proposed to factorize the latent space into multimodal discriminative factors and modality-specific generative factors. \citet{NEURIPS2020_43bb733c} also used the modality-specific code to improve the performance. Our analysis suggests that latent space factorization is necessary in the multimodal setting when the decoder is limited to the restricted model class.
% Our analysis necessitates the use of the modality specific code
% the consequence of align different encoder distribution is that we should avoid PoE and need to use modality specific code.

\section{Conclusion}
\label{sec:cons}
In this work, we proposed a new framework for multimodal generative modeling based on generative adversarial networks. To learn a coherent multimodal generative model, all unimodal encoder distributions were required to be aligned with the joint decoder distribution, simultaneously. The consequence of aligning multiple encoder distributions is that the product of experts should be avoided and the latent space need to be factorized. By factorizing the discriminator, we can utilize the unpaired data more efficiently and embed the constrastive learning mechanism in our framework. We conducted extensive experiments, whose promising results show that our proposed method outperforms the state-of-the-art method on a variety of metrics across different datasets. 

Despite of the encouraging result, we discuss the limitations of our approach. We found the performance drop if only more unimodal data are added while keeping the paired multimodal data unchanged. For our future work, we will investigate the semi-supervised learning and extend the method to the sequence generation on more challenging datasets.

%Our formulation suggests that feed more unimodal data to the factorized discriminators can improve density ratio estimation, which in turn will hopefully improve the encoders and decoders, but we found the performance degrades if only more unimodal data are added when keep paired multimodal data unchanged. The future work of this approach is to investigate semi-supervised learning and apply the method to sequence generation on more challenging datasets.
% \nocite{*}
\section*{Acknowledgments}
This work is supported by the National Natural Science Foundation of China under Grants (61831015).
\bibliography{aaai22}
\end{document}